\documentclass[conference]{IEEEtran}
\usepackage{cite}
\usepackage{placeins}
\usepackage{amsmath,amssymb,amsfonts}
\usepackage{algorithmic}
\usepackage{graphicx}
\usepackage{textcomp}
\usepackage{xcolor}
\usepackage{multirow}
\usepackage{booktabs}
\UseRawInputEncoding
\def\BibTeX{{\rm B\kern-.05em{\sc i\kern-.025em b}\kern-.08em
    T\kern-.1667em\lower.7ex\hbox{E}\kern-.125emX}}
\begin{document}

\title{Modular Deep Learning for Multivariate Time-Series: Decoupling Imputation and Downstream Tasks}

\author{\IEEEauthorblockN{1\textsuperscript{st} Joseph Arul Raj}
\IEEEauthorblockA{\textit{Biostatistics and Health Informatics} \\
\textit{King's College London}\\
London, United Kingdom \\
joseph.arulraj@kcl.ac.uk}
\and
\IEEEauthorblockN{2\textsuperscript{nd} Linglong Qian}
\IEEEauthorblockA{\textit{Biostatistics and Health Informatics} \\
\textit{King's College London}\\
London, United Kingdom \\
linglong.qian@kcl.ac.uk}
\and
\IEEEauthorblockN{3\textsuperscript{rd} Zina Ibrahim}
\IEEEauthorblockA{\textit{Biostatistics and Health Informatics} \\
\textit{King's College London}\\
London, United Kingdom \\
zina.ibrahim@kcl.ac.uk}
}

\maketitle 
\begin{abstract}
Missing values are pervasive in large-scale time-series data, posing challenges for reliable analysis and decision-making. Many neural architectures have been designed to model and impute the complex and heterogeneous missingness patterns of such data. Most existing methods are end-to-end, rendering imputation tightly coupled with downstream predictive tasks and leading to limited reusability of the trained model, reduced interpretability, and challenges in assessing model quality. In this paper, we call for a modular approach that decouples imputation and downstream tasks, enabling independent optimisation and greater adaptability. Using the largest open-source Python library for deep learning-based time-series analysis, PyPOTS, we evaluate a modular pipeline across six state-of-the-art models that perform imputation and prediction on seven datasets spanning multiple domains. Our results show that a modular approach maintains high performance while prioritising flexibility and reusability - qualities that are crucial for real-world applications. Through this work, we aim to demonstrate how modularity can benefit multivariate time-series analysis, achieving a balance between performance and adaptability.
\end{abstract}

\section{Introduction}
Modern data-driven systems increasingly rely on large and complex multivariate time-series data collected from sources such as longitudinal databases, wearable devices, sensors, and monitoring systems \cite{missingehr}. A persistent challenge in leveraging real-world data for predictive modeling and decision support is its inherent incompleteness, which arises naturally from irregular sampling, device failures, domain-specific recording patterns, and resource constraints \cite{missing}. In healthcare, missingness often reflects clinical recording practices \cite{howdeep}, while in sensor networks, it may result from sensor malfunctions or transmission loss \cite{sensornetworks}. Regardless of the origin, effective handling of missing values is essential for reliable inference. Traditional statistical imputation methods, while widely adopted, often fail to capture the complex temporal and spatial dependencies in real-world multivariate time series. Deep learning approaches have emerged as promising solutions by learning latent patterns within the incomplete data. While such architectures may achieve high predictive accuracy, current methodologies often integrate imputation and downstream tasks into monolithic, end-to-end pipelines, greatly reducing modularity and reusability. Such architectures suffer from three fundamental limitations:

\begin{itemize}
    \item \textbf{Reusability:} Adapting an end-to-end model to new downstream tasks requires retraining the entire system.
    \item \textbf{Attribution:} Improvements in downstream performance cannot be conclusively linked to the imputation model, the downstream architecture, or their interactions.
    \item \textbf{Maintainability:} Imputation error propagation to downstream training complicates debugging and optimisation.
\end{itemize}

To address these challenges, we investigate the strengths of a modular pipeline that decouples imputation from the downstream task. We experiment this by freezing the imputation backbone and training only the downstream task-specific head. Inspired by representation learning, we utilise pretrained imputation models to extract robust latent features from incomplete time-series, enabling a '\textit{train once, reuse many' paradigm}. Our study makes the following contributions:
\begin{itemize}
    \item \textbf{Benchmarking Modular vs. End-to-End:} We provide a comprehensive evaluation of six state-of-the-art models across seven datasets spanning three application domains, demonstrating that modular pipelines can achieve competitive performance against monolithic baselines.
    \item \textbf{Evaluation of Downstream Flexibility:} We empirically analyse the adaptability of pretrained imputers, showing that high-quality latent representations allow for the use of lightweight, interpretable downstream classifiers.
    \item \textbf{Representation Reusability Assessment:} We demonstrate the transferability of the decoupled models, highlighting utility in data-scarce regimes and varied tasks.
    \item \textbf{Efficiency Analysis:} We quantify the benefits of modularity in resource management, highlighting reductions in retraining costs for multi-task environments.
\end{itemize}

\begin{table*}[htbp]
\small
\centering
\renewcommand{\arraystretch}{0.85}
\caption{Imputation and Classification performances. 
\textbf{C:} Classification, \textbf{R:} Regression, \textbf{F:} Forecasting, \textbf{CRPS:} Continuous Ranked Probability Score, \textbf{CER:} Classification Error, \textbf{MMD:}  maximum Mean Discrepancy . Metrics are reported as presented in the original publications.}
\resizebox{\linewidth}{!}{
\begin{tabular}{|c|c|c|c|c|c|}
\hline
\textbf{Model} & \textbf{Architecture} & \textbf{Datasets} & \textbf{Imputation Performance} & \textbf{Tasks} & \textbf{Downstream Performance} \\ 
\hline
\multirow{2}{*}{GRU-D\cite{grud}} & \multirow{2}{*}{RNN} & Physionet & - & \multirow{2}{*}{C} & 0.842 (AUCROC) \\ \cline{3-4} \cline{6-6} 
& & MIMIC III & - & & 0.852 (AUCROC) \\ 
\hline
\multirow{3}{*}{BRITS\cite{brits}} & \multirow{3}{*}{RNN} & Physionet & 0.28 (MRE) & \multirow{3}{*}{C} & 0.85 (AUCROC) \\ \cline{3-4} \cline{6-6} 
& & Airquality & 11.56 (MRE) & & - \\ \cline{3-4} \cline{6-6} 
& & UCI localization & - & & 0.969 (AUCROC) \\
\hline
\multirow{4}{*}{SAITS\cite{saits}} & \multirow{4}{*}{Attention} & Physionet & 0.431(RMSE), 26.60\%(MRE) & \multirow{4}{*}{C} & 0.842 (AUCROC) \\ \cline{3-4} \cline{6-6}
& & Airquality & 0.518(RMSE), 19.30\%(MRE) & & - \\ \cline{3-4} \cline{6-6}
& & Electricity & 1.162(RMSE), 39.40\%(MRE) & & - \\ \cline{3-4} \cline{6-6}
& & ETT & 0.139(RMSE), 8.80\%(MRE) & & - \\
\hline
\multirow{3}{*}{TSI-GNN\cite{tsignn}} & \multirow{3}{*}{GNN} & Stocks & 0.029(30\%), 0.083(60\%) & \multirow{3}{*}{R} & 2.0653 ($R^2$) \\ \cline{3-4} \cline{6-6}
& & Energy & 0.136(30\%), 0.144(60\%) & & 3.127 ($R^2$) \\ \cline{3-4} \cline{6-6}
& & ICU & 0.074(30\%), 0.076(60\%) & & 1.01 ($R^2$) \\
\hline
\multirow{6}{*}{HI-VAE\cite{hivae}} & \multirow{6}{*}{MLP} & Adult & - & \multirow{6}{*}{C} & - \\ \cline{3-4} \cline{6-6}
& & Breast & - & & 0.052 (CER) \\ \cline{3-4} \cline{6-6}
& & Default credit & - & & 0.205 (CER) \\ \cline{3-4} \cline{6-6}
& & Letter & - & & 0.589 (CER) \\ \cline{3-4} \cline{6-6}
& & Spam & - & & 0.138 (CER) \\ \cline{3-4} \cline{6-6}
& & Wine & - & & 0.042 (CER) \\ 
\hline
\multirow{2}{*}{GP-VAE\cite{gpvae}} & \multirow{2}{*}{CNN} & Healing MNIST & 0.114(MSE) & \multirow{2}{*}{C} & - \\ \cline{3-4} \cline{6-6}
& & SPRITES & 0.002(MSE) & & 0.96 (AUCROC) \\
\hline
\multirow{2}{*}{V-RIN\cite{vrin}} & \multirow{2}{*}{RNN} & Physionet & 0.38(MSE), 0.407(MRE) & \multirow{2}{*}{C} & 0.84 (AUCROC) \\ \cline{3-4} \cline{6-6}
& & MIMIC  III & 0.586(MSE), 0.52(MRE) & & 0.86 (AUCROC) \\
\hline
\multirow{3}{*}{Supnot-MIWAE\cite{supnotmiwae}} & \multirow{3}{*}{\shortstack{CNN, \\ Attention}} & MIMIC III & 0.451(MRE), 0.149(MAE) & \multirow{3}{*}{C} & 0.86 (AUCROC) \\ \cline{3-4} \cline{6-6}
& & Human Activity & 0.373(MRE), 0.297(MAE) & & 0.883 (AUCROC) \\ \cline{3-4} \cline{6-6}
& & PhysioNet & 0.526(MRE), 0.367(MAE) & & 0.874 (AUCROC) \\
\hline
\multirow{2}{*}{GRUI-GAN\cite{gruigan}} & \multirow{2}{*}{RNN} & PhysioNet & 0.78(MSE) & \multirow{2}{*}{C} & 0.86 (AUCROC) \\ \cline{3-4} \cline{6-6}
& & KDD cup & 0.837(MSE) & & - \\
\hline
\multirow{2}{*}{$E^2$GAN\cite{e2gan}} & \multirow{2}{*}{RNN} & PhysioNet & - & \multirow{2}{*}{C} & 0.87 (AUCROC) \\ \cline{3-4} \cline{6-6}
& & KDD cup & 0.523(MSE) & & - \\
\hline
\multirow{3}{*}{SSGAN\cite{ssgan}} & \multirow{3}{*}{RNN} & Human Activity & 0.600(RMSE) & \multirow{3}{*}{C} & - \\ \cline{3-4} \cline{6-6}
& & PhysioNet & 0.598(RMSE) & & 0.85 (AUCROC) \\ \cline{3-4} \cline{6-6}
& & KDD cup & 0.435(RMSE) & & - \\
\hline
\multirow{5}{*}{SIM-GAN\cite{simgan}} & \multirow{5}{*}{CNN} & Leukaemia & 0.461(RMSE) & \multirow{5}{*}{C} & 0.993 (AUCROC) \\ \cline{3-4} \cline{6-6}
& & Colon tumour & 0.0497(RMSE) & & 0.952 (AUCROC) \\ \cline{3-4} \cline{6-6} 
& & DLBCL & 0.069(RMSE) & & 0.921 (AUCROC) \\ \cline{3-4} \cline{6-6} 
& & Lung Cancer & 0.126(RMSE) & & 0.971 (AUCROC) \\ \cline{3-4} \cline{6-6} 
& & Prostate Cancer & 0.047(RMSE) & & 0.989 (AUCROC) \\ 
\hline
\multirow{3}{*}{VIGAN\cite{vigan}} & \multirow{3}{*}{CNN} & MNIST & 0.138(RMSE) & \multirow{3}{*}{C} & - \\ \cline{3-4} \cline{6-6}
& & Cocaine-Opioid & 3.84(RMSE) & & 0.8 (Accuracy) \\ \cline{3-4} \cline{6-6}
& & Alcohol-Cannabis & - & & 0.64 (Accuracy) \\
\hline
\multirow{7}{*}{CSDI\cite{csdi}} & \multirow{7}{*}{Attention} & PhysioNet & 0.481(MAE) & - & - \\ \cline{3-4} \cline{6-6}
& & Air Quality & 9.60(MAE) & - & - \\ \cline{3-4} \cline{6-6}
& & Solar & - & F & 0.298 (CRPS) \\ \cline{3-4} \cline{6-6}
& & Electricity & - & F & 0.017 (CRPS) \\ \cline{3-4} \cline{6-6}
& & Traffic & - & F & 0.02 (CRPS) \\ \cline{3-4} \cline{6-6}
& & Taxi & - & F & 0.123 (CRPS) \\ \cline{3-4} \cline{6-6}
& & Wiki & - & F & 0.047 (CRPS) \\
\hline
\multirow{4}{*}{SSSD\cite{sssd}} & \multirow{4}{*}{CNN} & ECG data & 0.0118(RMSE) & - & - \\ \cline{3-4} \cline{6-6}
& & MuJoCo & 1.9(MSE) & - & - \\ \cline{3-4} \cline{6-6}
& & ETTm1 & 0.554(RMSE) & F & 0.612 (MSE) \\ \cline{3-4} \cline{6-6}
& & Solar & - & F & 5.03e2 (MSE) \\
\hline

\multirow{3}{*}{CSBI\cite{csbi}} & \multirow{3}{*}{\shortstack{CNN, \\ Attention}} & Synthetic Data & - & \multirow{3}{*}{-} & - \\ \cline{3-4} \cline{6-6}
& & PhysioNet & 0.63(RMSE) & & - \\ \cline{3-4} \cline{6-6}
& & Air Quality & 19(RMSE) & & - \\ 
\hline
\multirow{3}{*}{DA-TASWDM\cite{taswdm}} & \multirow{3}{*}{CNN} & MIMIC III & 0.806(RMSE) & \multirow{3}{*}{C} & 0.85 (AUCROC) \\ \cline{3-4} \cline{6-6}
& & Physionet 2019 & 0.723(RMSE) & & 0.81 (AUCROC) \\ \cline{3-4} \cline{6-6} 
& & Physionet & 0.698(RMSE) & & 0.83 (AUCROC) \\
\hline
\multirow{4}{*}{CSDE\cite{csde}} & \multirow{4}{*}{MLP} & PhysioNet & 0.728(MSE) & C & 1.057 (MSE) \\ \cline{3-4} \cline{6-6}
& & Speech Cmds & 0.423(MSE) & - & - \\ \cline{3-4} \cline{6-6}
& & Beijing Air & - & R & 1.671 (MSE) \\ \cline{3-4} \cline{6-6}
& & S\&P500 & - & R & 94.79 (MMD) \\
\hline
\end{tabular}
}
\label{tab:related_work_part1}
\end{table*}

\section{Related Work}
Table \ref{tab:related_work_part1} summarises the highest-performing deep learning models designed for time-series imputation and analysis. Examining the table, we identify the following critical gaps in the design and evaluation of existing imputation models: 
\begin{itemize} 
\item \textbf{Uncertain Improvement Source:} Many models report downstream performance only without imputation performance (e.g. GRU-D, HI-VAE, CDNet in the table). This implies that the systems were trained specifically for the downstream task, treating imputation as an internal "black box" mechanism rather than a distinct module. Such a design makes it unclear whether the reported performance is attributable to the strength of the imputation model, the complexity of the downstream classifier, or their interaction,  making targeted improvements difficult.

\item \textbf{Task Specificity:} Most models are evaluated on a single task per dataset, leaving the model's ability to accommodate different tasks common in the domain unknown. More importantly, many of those models, including GRU-D and GRUI-GAN, are monolithic joint learning models, updating the imputation weights based on the classification error propagation through the network, resulting in the limited task scope demonstrated in the table.

\item \textbf{Variability Across Publications:} There are no standardised models or benchmarks for comparing model outcomes, leading to varying reported performances across different publications. The reported performance of the most highly-cited models (e.g., BRITS \cite{brits}, SAITS \cite{saits}) appears to vary by publication.  For instance, even when using the same dataset, discrepancies arise in reported performance. In their publication, BRITS \cite{brits} reports an AUROC of 0.85 for downstream classification on the PhysioNet dataset. In \cite{saits}, SAITS claims an AUROC of 0.84, stating it is superior to BRITS, which they report as having an AUROC of 0.83. These inconsistencies make it challenging to establish clear benchmarks and hinder objective model comparison.

\end{itemize}

In contrast, modular architectures decouple imputation from prediction. This separation addresses the limitations of "black box" end-to-end models by ensuring that the imputation component learns valid data representations rather than overfitting to a specific downstream label. Prior work on ML pipelines emphasises that such independently trainable modules are more robust, consistent, and scalable; improving a single component (e.g., the imputer) benefits the overall system without requiring full retraining \cite{modi2023towards, sun2023modularity}.

While under-explored in time-series imputation, modular designs have proven effective in other high-stakes domains. In neuroimaging, decoupled pipelines for segmentation and classification have demonstrated superior clinical stratification compared to end-to-end counterparts \cite{yu2025end}. Similarly, in Large Language Models, decomposing architectures into distinct representation and reasoning modules has been found to mitigate hallucinations and improve interpretability \cite{wang2025modular}.

Furthermore, modularity offers distinct advantages for handling heterogeneous data. Independent modules prevent "module collapse," where a network fails to exploit its specialised components when optimised solely on a final loss function \cite{patil2025when, mittal2022modular}. This independence facilitates cross-domain transfer; for example, \cite{burlea2025modular} demonstrated that a modular pipeline generalised robustly across diverse dental datasets (radiographic, microbiome, transcriptome) where monolithic models struggled \cite{burlea2025modular}. Finally, a modular approach unlocks the potential of transfer learning. Pretrained feature extractors have been shown to outperform domain-specific deep networks in label-scarce environments \cite{gupta2020transfer}, validating the hypothesis that a robust, reusable imputer can compensate for simpler downstream classifiers.

\section{Methodology}
\subsection{Dataset}
To benchmark a modular pipeline against existing models, we used eight widely adopted time-series datasets for classification and regression tasks.  The datasets used span the healthcare, traffic, environment and electricity domains and exhibit diverse characteristics with respect to dimensionality, baseline missingness and feature types. The datasets are listed below, and their characteristics are shown in Table \ref{tab:datasets}.  As seen in the table, the datasets vary in domains, size, missingness distributions and feature types. 

\begin{itemize}
\item \textbf{Physionet 2012\cite{physionet}:} Extracted from the MIMIC-II database as part of the Computing in Cardiology Challenge 2012, it comprises records from 12,000 ICU stays of adult patients lasting at least 48 hours and is used for mortality classification. 
\item \textbf{eICU\cite{eicu}:} The eICU Collaborative Research Database is a multi-centre ICU database with high-granularity data for over 200,000 patient stays across the US in 2014-2015, including vital signs, diagnoses, and treatments; suitable for classification tasks such as mortality prediction. 
\item \textbf{MIMIC 89 \cite{mimic}:} A benchmark subset of the MIMIC-III database, used for mortality classification in critical care settings. 
\item \textbf{Physionet 2019\cite{physionet19}:} The PhysioNet Challenge 2019 dataset contains clinical data from ICU patients across three hospitals, with vital signs and lab values; designed for binary classification of early sepsis prediction. 
\item \textbf{Beijing Air Quality\cite{beijingair}:} A multivariate dataset tracking hourly air pollutant levels (e.g., PM2.5, O3) from 2013-2017 across Beijing stations, suitable for regression tasks like pollution forecasting. 
\item \textbf{Italy Air Quality\cite{italyair}:} Hourly air quality measurements from Italian sensors (2004-2005), including pollutants like CO and NO2, ideal for regression in environmental monitoring. 
\item \textbf{PeMS Traffic\cite{pems}:} Real-time traffic flow data from California highways (PeMS system), with sensor readings for speed and occupancy, used for regression in traffic prediction. 
\item \textbf{ETT (Electricity Transformer Temperature)\cite{ett}:} Hourly electricity load and oil temperature data from transformers (2016-2018), applied to regression for energy consumption prediction.
    
\end{itemize}

\begin{table}[t]
\centering
\caption{Dataset Characteristics: \textbf{Size:}samples $\times$ time steps $\times$ features. \textbf{Missing:} Average baseline missingness of all features. \textbf{Feature correlation:} average feature-wise Pearson correlation coefficient. \textbf{(static; categorical)} number of static and categorical features.}
\small
\resizebox{\columnwidth}{!}{%
\begin{tabular}{l@{\hskip 4pt}c@{\hskip 4pt}c@{\hskip 4pt}c@{\hskip 4pt}c@{\hskip 4pt}c}
\toprule
\textbf{Dataset} & \textbf{Domain} & \textbf{Size} & \textbf{Missing} & \textbf{Corr.} & \textbf{(Static; Cat.)} \\
\midrule
eICU         & Healthcare & $30680 \times 48 \times 20$ & 52.88\% & 0.11 & (6; 3)  \\
MIMIC\_89    & Healthcare & $14365 \times 48 \times 89$ & 97.16\% & 0.03 & (0; 0)\\
Physionet\_12   & Healthcare & $3997 \times 48  \times 35$ & 80.52\% & 0.06 & (0; 0) \\
Physionet\_19   & Healthcare & $4927 \times 48 \times 33$ & 78.38\% & 0.07 & (0; 0) \\
PEMS-BAY	    & Traffic & $731 \times 24 \times 862$ & 9.96\% & 0.56 & (0; 0)  \\
Italy\_air      & Environment & $777 \times 12 \times 13$ & 10.00\% & 0.44 & (0; 0)\\
Beiing\_air   & Environment & $1461 \times 24 \times 132$ & 11.41\% & 0.32 & (0; 0) \\
ETT         & Energy & $361 \times 48 \times 7$ & 9.85\% & 0.33 & (0; 0) \\
\bottomrule
\end{tabular} %
}
\label{tab:datasets}
\end{table}

\subsection{Experimental Environment}
We conduct all our experimental evaluation using PyPOTS \cite{pypots}, a comprehensive hub for deep learning-based imputation algorithms and the only available open-source Python library for handling incomplete multivariate time series. PyPOTS provides a standardised and reproducible environment for time-series imputation, classification, clustering, and forecasting. The PyPOTS ecosystem is equipped with complementary libraries that facilitate benchmarking \cite{pypots}. Those include,\textbf{TSDB (Time Series Data Beans)}, a data loader with access to hundreds of publicly available datasets, \textbf{BenchPOTS}, which offers standardised preprocessing pipelines and evaluation protocols, and \textbf{PyGrinder}, which simulates the real-world missingness for ground-truth evaluation. PyGrinder supports different missing data mechanisms, namely MCAR (missing completely at random), MAR (missing at random), and MNAR (missing not at random), ensuring that experiments can mimic a variety of real-world scenarios.

\subsection{Models Evaluated }
The selection of imputation models was guided by the need for architectural diversity to evaluate the modular framework's robustness while ensuring compatibility with PyPOTS for standardised implementation and reproducible evaluation across datasets. We prioritised models that excel in handling multivariate time-series missingness, with proven efficacy in both classification-heavy domains (e.g. medicine) as well as regression-heavy ones (e.g. environment/traffic). The models were grouped by their core architectures as: RNN-based, CNN-based, and Transformer-based models. This grouping allows us to assess which paradigms yield the most reusable representations for decoupled downstream tasks
\begin{itemize}
    \item\textbf{RNN-based Models (CSAI, BRITS, GRUD):} These models leverage recurrent structures to capture temporal dependencies, making them ideal for irregular time-series like medical data. \textbf{BRITS (Bidirectional Recurrent Imputation for Time Series)}\cite{brits}: Utilises bidirectional recurrent neural networks (RNNs) to impute missing values by combining two estimates: (i) temporal estimates from decayed historical states and (ii) feature-based estimates via fully connected layers. \textbf{CSAI (Conditional Self-Attention Imputation)}\cite{csai}: Extends BRITS with three EHR-focused enhancements: a) attention-based hidden state initialisation; b) clinical-pattern-aware temporal decay; and c) non-uniform masking to address non-random missingness. \textbf{GRU-D}\cite{grud}: A recurrent neural network based on the Gated Recurrent Unit (GRU) architecture, incorporating temporal decay mechanisms to address irregular time intervals in clinical data.
    \item \textbf{CNN-based Model (TimesNet):} A temporal 2D-variation network that transforms 1D time-series into 2D representations using multi-periodicity, enabling better capture of intra and inter-period variations for imputation and prediction.
    \item \textbf{Transformer-based Models (SAITS, Autoformer):} These models employ attention mechanisms for global dependency modelling, offering scalability for long sequences. \textbf{SAITS (Self-Attention Imputation for Time Series)}\cite{saits}: Uses diagonal masking and self-attention to learn missing values from observed data, focusing on joint modelling of features and temporal dynamics. \textbf{Autoformer}\cite{autoformer}: An attention-based model with decomposition architecture and auto-correlation mechanism, designed for long-range dependencies in time-series, adaptable for imputation in forecasting-heavy scenarios.
\end{itemize}
The RNN-based models were chosen for their established performance in classification tasks, particularly on medical datasets like Physionet and models like TimesNet and Autoformer were chosen for their strong forecasting abilities in regression scenarios.

\subsection{Training and Evaluation Rubric} \label{sec:rubric}

All datasets were obtained from the TSDB repository within the PyPOTS ecosystem. Synthetic missingness was introduced using PyGrinder. The processed datasets were then used to train the imputation models provided in PyPOTS in a supervised reconstruction setting, where the masks of the artificially removed data points served as the ground truth. Each model was trained with early stopping, and the checkpoint corresponding to the best validation performance (based on the lowest Mean Squared Error) was retained. The saved models were subsequently reloaded, and their weights were frozen to generate the hidden representations, which were then fed into the predictive heads corresponding to the downstream tasks.

Model performances were evaluated at these levels:
\begin{itemize}
\item \textbf{Performance:} using the Area Under the Receiver Operating Characteristic Curve (AUROC) for classification and Mean Absolute Error (MAE) for regression. 
\item \textbf{Time Efficiency:} measuring real-time inference latency of End-to-End (E2E) pipelines relative to modular baselines to quantify the computational overhead and measured using the Inference Latency Ratio to assess real-time deployment feasibility. Here, we report the normalised inference time of End-to-End (E2E) pipelines relative to a modular Single-layer MLP baseline. A ratio greater than $1.0$ indicates that the E2E approach introduces computational overhead.

\item \textbf{Data Efficiency:} In addition to evaluating model stability under a label-rich regime (the downstream predictor is trained on 100\% of the available training labels), we also evaluate performance on a data-scarce regime where the downstream predictor is trained on a random subset of 10\% of the available training data. The resulting metrics act as a proxy for scenarios where the label annotations are limited, highlighting the risk of overfitting in coupled architectures versus the stability of frozen, pre-trained representations.

\item \textbf{Transferability:} We evaluate the downstream performance of a model pretrained on a source dataset when applied to a different target dataset from the same domain without retraining, evaluating deployment without backbone retraining. 


\end{itemize}

\section{Experimental setup}

\subsection{Experimental Design}

\begin{figure}[ht!]
\centering
\includegraphics[width=\linewidth]{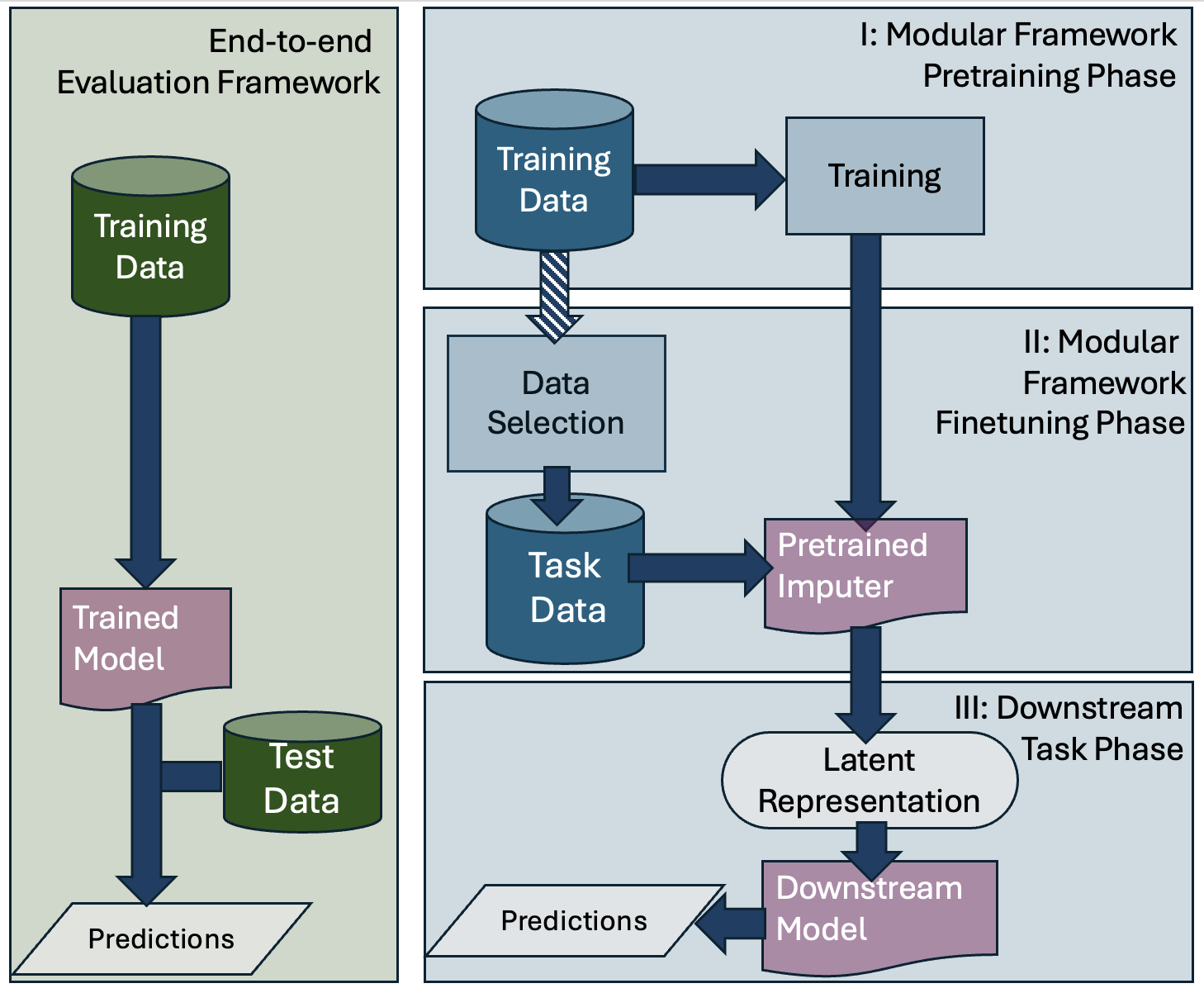}
\caption{The three phases of our modular pipeline.} \label{fig:framework}
\end{figure}

The diagram shown in Fig.\ref{fig:framework} details our structured approach we followed to compare the modular pipeline to an end-to-end framework according to the rubric outlines in Section \ref{sec:rubric}. While the end-to-end model follows the traditional pipeline, the modular approaches comprises three phases. In phase I, an imputation model is pretrained on a source dataset $\mathcal{D}_X$ with artificial missingness and in phase II, the pretrained imputer model processes task data $\mathcal{D}_T$ to produce latent representations. Here, $\mathcal{D}_T$ can be the full pretraining data $\mathcal{D}_X$, a 10\% subset of  $\mathcal{D}_X$  (data efficiency), or a different target dataset $\mathcal{D}_Y$(transferability). In Phase III, we have downstream models that can either be a classifier or a regressor to operate on the latent representation and generate final predictions.

\subsection{Downstream Tasks}

Each imputer’s latent representations (hidden states) were extracted according to its architecture, i.e, final hidden states for recurrent models, mean encoder outputs for attention-based models, and flattened encoder features for convolutional models. For bidirectional models, forward and backward states were concatenated before passing them to the downstream network. For the regression task, by taking a multivariate many-to-many forecasting task as the objective, the downstream heads were trained to simultaneously predict the values of all features at the final timestep ($T$) based on the latent history of the preceding sequence. 
To evaluate the modularity of the pretrained imputers, two downstream architectures of differing complexity were implemented: a single-layer MLP and a five-layer MLP, representing simple and complex predictive heads, respectively, were employed. Both architectures were trained for 200 epochs using the Adam optimiser and early stopping (patience = 10). Loss functions were task-specific: Dice BCE or cross-entropy for classification, and MSE for regression. 

Two downstream heads were implemented to test information retention: a \textbf{Simple Head} (Single-layer Linear) and a \textbf{Complex Head} (5-layer MLP with ReLU activations). All classification tasks utilised \textbf{Cross-Entropy Loss} to ensure direct comparability with the End-to-End (E2E) baselines.

\subsection{Implementation Details}
Using PyPOTS, datasets were sourced from \textbf{TSDB}, with artificial missingness simulated via \textbf{PyGrinder}. For all models except CSAI, synthetic missingness was introduced using PyGrinder with 10\% point-missingness. In the case of CSAI, data were regenerated with a 0\% missing rate in PyGrinder and subsequently masked using CSAI’s internal non-uniform random masking mechanism to achieve an equivalent 10\% overall missingness.

Each model was trained for 200 epochs. RNN-based models employed a hidden size of 1024, and the number of layers in both SAITS and Autoformer were restricted to one to maintain consistency and computational efficiency. Hyperparameters were optimised using PyPOTS’s benchmarking utility, which integrates an NNI-based AutoML procedure to tune learning rate, batch size, and other model-specific parameters.

All experiments were executed on an NVIDIA H100 GPU (94 GB memory). Minor modifications were made to the PyPOTS source code to enable downstream analysis: the original end-to-end pipeline was extended to return latent representations alongside the imputed sequences. For bidirectional models, both forward and backwards hidden states were concatenated to form the input representation for the downstream heads. Since these code modifications were incompatible with the released PyPOTS package, all experiments were conducted using a locally built version of the library.

\section{Experimental Results}
Our experimental results can be summarised in two categories: classification and regression tasks, described below.

\subsection{Classification}

Table \ref{tab:classification} presents the classification performance across four clinical datasets. Across the majority of model-dataset pairs, the modular approach consistently matches or outperforms the E2E baseline. For instance, on the PhysioNet12 dataset, modular configurations achieve the highest accuracy for all six evaluated models. Notably, increasing the complexity of the downstream classifier (from ‘S’ to ‘C’) does not guarantee better performance; in several cases, especially with large datasets such as eICU and MIMIC, the simpler Single-layer head (‘S’) yields superior results. This suggests that the latent representations learned by these self-supervised backbones are robust and linearly separable, often rendering deep, non-linear classifiers redundant.

Furthermore, End-to-End fine-tuning frequently leads to performance degradation. This is most visible with SAITS on PhysioNet19, where accuracy drops from 0.679 (Modular ‘C’) to 0.591 (E2E). While E2E training offers marginal gains in specific high-data regimes such as GRU-D on MIMIC89 and eICU, it lacks the consistency of the modular approach, which provides stable, high-performance results without the need for computationally expensive retraining.

\begin{table}[!ht]
\centering
\caption{Classification performance : 
\textbf{S} = Single-layer MLP, \textbf{C} = Complex 5-layer MLP, \textbf{E2E} = End-to-End performance through PyPots; Best values per row–dataset group are in bold.}
\resizebox{\columnwidth}{!}{\large 
\begin{tabular}{lcccccccccccc}
\toprule
 & \multicolumn{3}{c}{PhysioNet12} & \multicolumn{3}{c}{PhysioNet19} & \multicolumn{3}{c}{MIMIC89} & \multicolumn{3}{c}{eICU}  \\
\cmidrule(lr){2-4} \cmidrule(lr){5-7} \cmidrule(lr){8-10} \cmidrule(lr){11-13}
\textbf{Model} & S & C & E2E & S & C & E2E & S & C & E2E & S & C & E2E   \\
\midrule
GRU-D      & \textbf{0.839}& 0.835& 0.824& 0.669& \textbf{0.734}& 0.713& 0.835& 0.827& \textbf{0.848}& 0.881& 0.879& \textbf{0.882}\\
BRITS      & 0.837& \textbf{0.845}& 0.808& \textbf{0.698}& 0.675& 0.616& 0.825& \textbf{0.838}& 0.780& \textbf{0.880}& \textbf{0.880}& 0.830\\
SAITS      & 0.861& \textbf{0.864}& 0.848& 0.676& \textbf{0.679}& 0.591& \textbf{0.843}& 0.838& 0.824& \textbf{0.886}& 0.885& 0.884\\
CSAI       & 0.836& \textbf{0.850}& 0.847& \textbf{0.680}& 0.673& 0.669& \textbf{0.834}& 0.823& 0.827& \textbf{0.890}& \textbf{0.890}& 0.887\\
TimesNet   & \textbf{0.829}& 0.816& 0.820& \textbf{0.660}& 0.637& 0.579& \textbf{0.798}& 0.785& 0.796& 0.844& 0.844& \textbf{0.845}\\
Autoformer & \textbf{0.784}& 0.759& 0.779& 0.671& \textbf{0.769}& 0.747& \textbf{0.813}& 0.798& 0.807& \textbf{0.831}& 0.815& 0.799\\
\bottomrule
\end{tabular}
}
\label{tab:classification}
\end{table}

\subsection{Regression}

In contrast to classification, the regression results in Table \ref{tab:regression} demonstrate a strong uniform preference for the model depth. For every model across all four datasets (BeijingAir, ItalyAir, ETT, and Traffic), the Complex 5-layer MLP (‘C’) significantly outperforms the Single-layer baseline (‘S’). The performance gap is often substantial. For example, using Autoformer on the BeijingAir dataset, the error reduces from 0.338 (‘S’) to 0.250 (‘C’). Similarly, GRU-D on ItalyAir sees an improvement from 0.379 to 0.299. 
\begin{table}[!ht]
\centering
\caption{Regression performance : 
\textbf{S} = Single-layer MLP, \textbf{C} = Complex 5-layer MLP. Lower values are better; 
\textbf{E2E} = End-to-End performance through PyPots. Best value per model–dataset group in bold.
\textbf{NA} = Model not implemented in PyPOTS}
\resizebox{\columnwidth}{!}{\Large
\begin{tabular}{lcccccccccccc}
\toprule
 & \multicolumn{3}{c}{BeijingAir} & \multicolumn{3}{c}{ItalyAir} & \multicolumn{3}{c}{ETT} & \multicolumn{3}{c}{Traffic} \\
\cmidrule(lr){2-4} \cmidrule(lr){5-7} \cmidrule(lr){8-10} \cmidrule(lr){11-13} 
\textbf{Model} & S & C & E2E & S & C & E2E & S & C & E2E & S & C & E2E   \\
\midrule
GRU-D      & 0.261 & \textbf{0.239} & NA & 0.379 & \textbf{0.299} & NA & 0.339 & \textbf{0.334} & NA & 0.225 & \textbf{0.194} & NA \\
BRITS      & 0.249 & \textbf{0.233} & NA & 0.355 & \textbf{0.332} & NA & 0.341 & \textbf{0.328} & NA & 0.220 & \textbf{0.193} & NA \\
SAITS      & 0.304 & \textbf{0.267} & NA & 0.380 & \textbf{0.336} & NA & 0.359 & \textbf{0.346} & NA & 0.263 & \textbf{0.262} & NA \\
CSAI       & 0.275 & \textbf{0.237} & NA & 0.366 & \textbf{0.289} & NA & 0.355 & \textbf{0.341} & NA & 0.233 & \textbf{0.207} & NA \\
TimesNet   & 0.309 & \textbf{0.284} & NA & 0.509 & \textbf{0.350} & NA & 0.351 & \textbf{0.329} & NA & 0.233 & \textbf{0.212} & NA \\
Autoformer & 0.338 & \textbf{0.250} & 0.294 & 0.381 & \textbf{0.285} & 0.478 & 0.390 & \textbf{0.338} & 0.413 & 0.245 & 0.210 & \textbf{0.153} \\
\bottomrule
\end{tabular}
}
\label{tab:regression}
\end{table}

Data for E2E regression performance is limited only to the Autoformer model because these architectures were originally proposed and implemented with classification or imputation objectives in mind. This limitation unintentionally highlights a primary advantage of the modular paradigm, \textbf{task flexibility}. By decoupling the backbone, we successfully repurposed these classification-only models for regression simply by attaching a continuous output head capability that the rigidly coupled E2E frameworks lack. The available data from Autoformer in the E2E setting shows that, while Autoformer E2E achieves the best overall score on the Traffic dataset (0.153), it performs significantly worse than the modular baselines on the other three datasets(BeijingAir, ItalyAir, ETT). This instability further reinforces the reliability of the modular approach, where performance gains can be systematically achieved simply by scaling the downstream head.

\section{Ablation Study}
\subsection{Computational Cost of E2E Coupling}
\begin{figure}[h!]
\includegraphics[width=\linewidth]{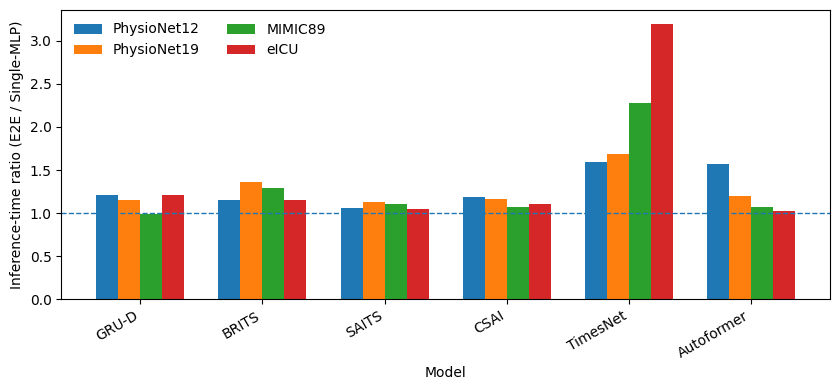}
\caption{Normalised inference-time overhead of end-to-end (E2E) architectures compared to modular baselines. The y-axis represents the ratio of E2E inference time to that of a single-layer modular MLP, with the dashed line at $1.0$ indicating parity.} \label{fig:time}
\end{figure}
The diagram shown in Fig.\ref{fig:time} illustrates the inference time overhead of end-to-end (E2E) pipelines compared to a single-layer modular classifier across various models and datasets. Each bar represents the ratio between E2E and modular inference time (E2E / Single-MLP), with the dashed horizontal line indicating parity. Values above this threshold correspond to slower inference under E2E coupling. \\
The graph reveals two critical trends supporting the modular approach. First, E2E integration offers no efficiency upside. Across all evaluated models, ratios consistently sit at or above the $1.0$ parity line. Even for a lightweight backbone like GRU-D, the E2E pipeline fails to outperform the modular baseline. This indicates that the theoretical seamlessness of end-to-end processing does not translate into faster inference; at best, it almost matches the speed of a decoupled system. Second, coupled architectures scale poorly with backbone complexity. This fragility is most evident in the TimesNet and Autoformer results. In a modular framework, the heavy computational cost of these backbones (e.g., 2D convolutions in TimesNet) is incurred only once during feature extraction. In the E2E paradigm, this cost is paid on every forward pass. The severe latency spikes were observed in TimesNet, particularly on the larger eICU dataset ($>3.0\times$ baseline). Overall, the figure highlights that while E2E coupling may appear architecturally appealing, its inference-time cost is strongly backbone dependent as E2E requires full gradient calculation through the temporal backbone, whereas the modular approach only requires optimising the lightweight MLP head.

\subsection{Low-Resource Training}
To simulate a real-world clinical scenario, where unlabeled patient data is abundant, but the annotated labels are scarce, we evaluated classification performance using only 10\% of the available training labels, while validation and testing were conducted on the full dataset to ensure statistical significance (Table \ref{tab:classification_10}). This low-resource training exposes a critical fragility in End-to-End (E2E) models. \\
From Table \ref{tab:classification_10}, the most immediate observation is the catastrophic degradation in performance of E2E models on smaller datasets. On PhysioNet12 ($N=255$ unique training samples), transformer-based models such as SAITS and CSAI show performances closer to random guessing under E2E training (0.464 and 0.465, respectively). This indicates that when labelled data is insufficient, attempting to update the high-dimensional backbone parameters leads to severe overfitting. Even on the larger eICU dataset, these complex E2E models fail to recover, with CSAI dropping from 0.890 (full data) to 0.625 (10\% data). The E2E approach effectively demands a large volume of labels to stabilise the optimisation of the full pipeline. 
In contrast, modular architectures demonstrate stability \textbf{in label-scarce settings}. By freezing the backbone, pre-trained on the full unlabelled corpus and training only a lightweight head, the system retains robust feature extraction. For example, the modular CSAI model with a Complex head (‘C’) achieves an AUC of 0.832 on PhysioNet12, nearly matching its full-data performance (0.850) and significantly outperforming its E2E counterpart (0.465)

Furthermore, we observe a shift in the optimal downstream head complexity. Unlike the full-data regime, where the single-layer perceptron (‘S’) was often sufficient, the 10\% setting frequently favours the Complex 5-layer head (‘C’). For instance, on the eICU dataset, the Complex head outperforms the Single head across four of the six models. This suggests that when the backbone is frozen and label density is low, a slightly more expressive classifier is better equipped to find a robust decision boundary with the pre-learned feature space, without the risk of degradation inherent to E2E fine-tuning.

\begin{table}[!ht]
\centering
\caption{Classification performance with only 10\% of the training data. Values in parentheses denote the total number of unique patient records.
\textbf{S} = Single-layer MLP, \textbf{C} = Complex 5-layer MLP, \textbf{E2E} = End-to-End performance through PyPots; Best values per row–dataset group are in bold.}
\resizebox{\columnwidth}{!}{\large 
\begin{tabular}{lccccccccc}
\toprule
 & \multicolumn{3}{c}{PhysioNet12(255)} &  \multicolumn{3}{c}{MIMIC89(861)} & \multicolumn{3}{c}{eICU(920)}  \\
\cmidrule(lr){2-4} \cmidrule(lr){5-7} \cmidrule(lr){8-10} 
\textbf{Model} & S & C & E2E & S & C & E2E & S & C & E2E   \\
\midrule
GRU-D      & 0.723& \textbf{0.779}& 0.743& \textbf{0.774}& 0.731& 0.752& 0.832& 0.834& \textbf{0.838}\\
BRITS      & 0.695& \textbf{0.727}& 0.643& 0.724& 0.775& \textbf{0.789}& \textbf{0.817}& 0.809& 0.816\\
SAITS      & \textbf{0.777}& 0.690& 0.464& \textbf{0.759}& 0.745& 0.596& 0.815& \textbf{0.830}& 0.652\\
CSAI       & 0.772& \textbf{0.832}& 0.465& 0.759& \textbf{0.779}& 0.756& 0.840& \textbf{0.857}& 0.625\\
TimesNet   & \textbf{0.725}& 0.694& 0.507& 0.743& \textbf{0.773}& 0.721& 0.787& \textbf{0.808}& 0.757\\
Autoformer & \textbf{0.676}& 0.625& 0.558& 0.713& \textbf{0.718}& 0.585& 0.749& \textbf{0.766}& 0.685\\
\bottomrule
\end{tabular}
}
\label{tab:classification_10}
\end{table}

\subsection{Transferability}
Given the consistently challenging performance on the PhysioNet19 dataset (baseline AUROC $<0.70$ for almost all models), we investigated the potential of transfer learning. We replaced the standard backbone, which was pretrained on PhysioNet19, with backbones pretrained on external, larger-scale clinical datasets (eICU and MIMIC89). We then trained a modular Single-layer MLP on the PhysioNet19 target labels while keeping these external backbones frozen. 

As shown in Table \ref{tab:transfer_learning}, the modular architecture successfully facilitates cross-dataset transfer without huge degradation in performance. Notably, transferring representations from the PhysioNet2012 proved highly effective, surpassing the native baselines. For instance, the CSAI model, with the PhysioNet2012 backbone, improved performance from $0.680$ (native) to $0.71$ (transfer) on the simple MLP, with even higher gains using the complex MLP ($0.741$). While GRU-D results were more variable, the experiment validates a crucial architectural advantage: \textbf{interoperability}. In a modular framework, a high-quality backbone can be trained once on a source dataset and deployed across multiple downstream tasks with minimal computational overhead. This capability is inherently difficult to achieve in End-to-End pipelines, which typically require joint optimisation on the specific target distribution.

\begin{table}[!ht]
\centering
\caption{Transfer learning performance on the \textbf{PhysioNet19} dataset. 
\textbf{Native}: Backbone trained on PhysioNet2019 (Baseline). 
\textbf{Transfer}: Backbone trained on eICU or MIMIC89 or Physionet2012. 
All use a modular downstream classifier.}
\label{tab:transfer_learning}
\resizebox{0.8\columnwidth}{!}{
\begin{tabular}{lcccc}
\toprule
 & \multicolumn{1}{c}{\textbf{Native Source}} & \multicolumn{2}{c}{\textbf{External Source (Transfer)}} \\
\cmidrule(lr){2-2} \cmidrule(lr){3-5}
\textbf{Model} & PhysioNet19 & eICU & MIMIC89 & Physionet2012\\
\midrule
BRITS & 0.698/0.675 & 0.684/0.673 & 0.645/0.632 & \textbf{0.702/0.709}\\
CSAI  & 0.680/0.673 & 0.697/0.725 & 0.705/0.725 & \textbf{0.71/0.741} \\
GRUD  & \textbf{0.666}/0.734 & 0.565/0.672 & 0.456/\textbf{0.762} & 0.645/0.718 \\
\bottomrule
\midrule
\multicolumn{5}{l}{\footnotesize\textit{Values shown as Single-layer MLP (1MLP) / Complex MLP (5MLP)}} \\

\end{tabular}
}
\end{table}

\section{Conclusion}
In this work, we presented a modular deep learning framework for handling missing values in multivariate time series and downstream analysis, evaluating its performance across six state-of-the-art models and eight diverse datasets. Through this extensive evaluation, we demonstrate that decoupling imputation from downstream prediction offers a robust and efficient alternative to rigid E2E pipelines. Our results show that, when there is abundant labelled data, modular architectures perform just as well as fully end-to-end models. In the low-label setting, they tend to outperform the E2E models by more than 30 percentage points in AUC. Furthermore, our work also highlights the practical flexibility of the modular frameworks. By using the learned representation of the models originally designed for classification (such as Autoformer and TimesNet), we were able to repurpose those models for regression tasks. We have also demonstrated the ability to do transfer learning between different datasets, matching or even surpassing the original baseline performances. This approach of treating the imputer as a reusable feature extractor can reduce the marginal computational cost for new clinical tasks. In conclusion, we propose that modularity is Pareto-optimal as we can gain robustness in data scarcity, flexibility across tasks and domains and lower the computational cost without meaningful sacrifice in accuracy

 \bibliographystyle{IEEEtran}
 \bibliography{citations}

\end{document}